\documentclass{article}
\usepackage{spconf,amsmath,graphicx}
\usepackage{soul}
\usepackage{amsmath}
\usepackage{amssymb}
\usepackage{algorithmic}
\usepackage{graphicx}
\usepackage{textcomp}
\usepackage{xcolor}
\usepackage{algorithm} 
\usepackage{algorithmic}
\usepackage{multirow}
\usepackage{subfigure}
\usepackage{enumitem}

\newcommand{\methodname}{{\tt{FairFedJS}}}

\title{Fairness-Aware Job Scheduling for Multi-Job Federated Learning}
\name{Yuxin Shi$^{1,2,3}$, Han Yu$^{1}$
\thanks{This research is supported by National Research Foundation, Singapore and DSO National Laboratories under the AI Singapore Programme (AISG Award No: AISG2-RP-2020-019); Alibaba Group through Alibaba Innovative Research (AIR) Program and Alibaba-NTU Singapore Joint Research Institute (JRI), Nanyang Technological University, Singapore; and the RIE 2020 Advanced Manufacturing and Engineering (AME) Programmatic Fund (No. A20G8b0102), Singapore.}}

\address{$^{1}$School of Computer Science and Engineering, Nanyang Technological University (NTU), Singapore \\
$^{2}$Alibaba-NTU Singapore Joint Research Institute, NTU, Singapore.
$^{3}$Alibaba Group, Hangzhou, China \\
\{shiy0029, han.yu\}@ntu.edu.sg}
\begin{document}
%
\maketitle
\begin{abstract}
Federated learning (FL) enables multiple data owners (a.k.a. FL clients) to collaboratively train machine learning models without disclosing sensitive private data. Existing FL research mostly focuses on the monopoly scenario in which a single FL server selects a subset of FL clients to update their local models in each round of training. In practice, there can be multiple FL servers simultaneously trying to select clients from the same pool. In this paper, we propose a first-of-its-kind \underline{Fair}ness-aware \underline{Fed}erated \underline{J}ob \underline{S}cheduling (\methodname{}) approach to bridge this gap. Based on Lyapunov optimization, it ensures fair allocation of high-demand FL client datasets to FL jobs in need of them, by jointly considering the current demand and the job payment bids, in order to prevent prolonged waiting.
Extensive experiments comparing \methodname{} against four state-of-the-art approaches on two datasets demonstrate its significant advantages. It outperforms the best baseline by 31.9\% and 1.0\% on average in terms of scheduling fairness and convergence time, respectively, while achieving comparable test accuracy.

\end{abstract}
\section{Introduction}
\label{sec:intro}
The success of artificial intelligence (AI) hinges on the availability of sufficient data to train effective machine learning (ML) models~\cite{chen2022noise}. However, direct access to distributively owned datasets risks infringing on user privacy and violation of laws such as the General Data Protection Regulation~\cite{GDPR2017}. To this end, federated learning (FL)~\cite{kairouz2021advances,FL:2019} has emerged as an alternative collaborative ML paradigm that operates without the need for direct access to sensitive local data in order to address privacy concerns.
With FL come new challenges.
Existing FL research mostly focuses on the monopoly scenario in which one FL server selects data owners (a.k.a. FL clients) from a common pool to perform local model updates in each round of FL training~\cite{Nishio2019ClientSF, huang2020efficiency, shi2023fairness}. However, in practice, there can be multiple FL servers competing for FL clients at any given time~\cite{chen2023metafed}. 
The problem of FL job scheduling with multiple FL servers remains open~\cite{zhou2022efficient}. 

A limited number of works have emerged to explore this problem~\cite{zhou2022efficient, wang2023confederated, tan2023reputation}. Tan \textit{et al.}~\cite{tan2023reputation}
proposed a reputation-based approach to help multiple federations optimize the costs for recruiting clients without compromising utility. It focuses on dealing with multi-task FL in which the tasks proposed by multiple federations share some common parts of the model. Zhou \textit{et al.}~\cite{zhou2022efficient} introduced a new concept called Multi-job FL, which deals with the simultaneous training of multiple independent jobs without exchanging models across jobs~\cite{liu2022multi}. Since one client is assumed to be able to accept only one job at a time, \cite{zhou2022efficient} proposed two scheduling methods to assign clients to different FL jobs, while minimizing costs.

However, these approaches did not consider the scheduling of the FL jobs. In multi-job FL, clients possess diverse capabilities. This implies that each client can own multiple datasets (e.g., images, speech). Consequently, a single client can undertake multiple types of jobs. However, following the common assumption that each client can only handle one job at a time, competition might arise among FL servers for access to reliable clients. FL servers adjust their job payment bids to gain priority for client selection in such a competitive situation. On the other hand, clients with data types that are in high demand shall adapt the acceptance of jobs to optimize their benefits. Nevertheless, there is a lack of decision-support methods to efficiently manage this process.

In this paper, we bridge this important gap by proposing the \underline{Fair}ness-aware \underline{Fed}erated \underline{J}ob \underline{S}cheduling (\methodname{}) approach. 
Based on Lyapunov optimization \cite{Yu-et-al:2016,Yu-et-al:2017}, it ensures fair allocation of high-demand FL client datasets to FL jobs in need of them, by jointly considering the current demand and the job payment bids, in order to prevent prolonged waiting. 
It offers FL servers the flexibility to adapt their payments according to current demand and supply for FL jobs to achieve their desired ordering in the FL job schedules in a human-over-the-loop manner. 
Extensive experiments comparing \methodname{} against four state-of-the-art approaches on two real-world datasets demonstrate its significant advantages. Specifically, it achieves 31.9\% higher scheduling fairness and 1.0\% faster convergence on average compared to the best-performing baseline, with comparable test accuracy.
To the best of our knowledge, \methodname{} is the first fairness-aware job scheduling approach for multi-job FL.

\vspace*{-1mm}
\section{Preliminaries}
In this paper, we focus on an environment consisting of multiple FL servers (a.k.a., job publishers) and multiple FL clients. We introduce a multi-job FL framework consisting of two components: 1) a multi-job scheduler and 2) a payment model. These components help task publishers optimize their payments for future rounds of FL training.

\textbf{Settings}: We study an FL system involving $N$ FL clients and $K$ FL jobs published. Each client $i$ can only accept one job at a given time slot $t$. $i$ can own multiple datasets $D_{i,m}$ where $m$ indicates the data type (e.g., image, speech, search history). $i$ can indicate the costs $c_{i,m}$ for using its different datasets. Each FL job requires a type of data for model training. A job $k$ is published with the number of clients it requires, $n_{k,m}$, and its payment, $p_k(t)$, for recruiting them.


\textbf{Cost Model}: The average cost incurred for mobilizing a set of clients with data type $m$ is expressed as $\hat{c}_m(t) = \frac{1}{N_m} \sum_{i=1}^{N_m} c_{i,m}(t)$.
The average reliability value for clients with data type $m$ is expressed as
$\hat{r}_m(t) = \frac{1}{N_m} \sum_{i=1}^{N_m} r_{i,m}(t)$.
The cost for mobilizing a set of clients for a job $k$ is:
\begin{equation}
    c_k(t) = \sum_{m=1}^M \frac{\hat{c}_m(t)}{\hat{r}_m(t)} a_{k,m}(t).
\end{equation}
where $a_{k,m}(t) \geq 0$ is the supply of clients with dataset $m$ for job $k$.
The total revenue $f(t)$ for the FL system is linearly related to the collective payment for all jobs, defined as $f(t) = \sum_{k=1}^K \frac{a_k(t)}{n_k} p_k(t)$, 
where $a_k(t) = \sum_{m=1}^M a_{k,m}(t)$ and $n_k = \sum_{m=1}^M n_{k,m}$ when a job requires multiple datasets. Since we focus on the horizontal FL~\cite{FL:2019} setting where a job only requires one data type (i.e., $m=1$ for each job $k$), $a_k(t)$ and $n_k$ can be simplified as $a_k(t) = a_{k,m}(t)$ and $n_k = n_{k,m}$.

\textbf{FL Client Selection}:
For each job, clients are selected considering both their reputation values~\cite{yu2014reputation,yu2014filtering} and data fairness~\cite{Shi-et-al:2023TNNLS}. The criteria for selecting a client $i$ with dataset type $m$ for job $k$ is expressed as:
\begin{equation}
    \gamma_{i,k,m} = r_{i,m}(t) - \beta \mathcal{F}_{i,k,m}(t)
\end{equation}
where $r_{i,m}(t)$ is the reputation of client $i$ regarding its dataset of type $m$. $\mathcal{F}_{i,k,m}(t)$ is the fairness achieved by the system if client $i$ is selected to contribute its dataset of type $m$ to job $k$ at $t$. $\beta>0$ is a normalization factor. During client selection, a job aims to select clients with the highest $\gamma_{i,k,m}$ values.

The reputation model of choice is built upon the Beta Reputation System (BRS)~\cite{Jøsang02thebeta}. It manages a distinct client reputation table for each specific data type $m$. BRS calculates each client’s reputation based on its previous performance track records. The reputation score $r_{i,m} \in [0,1]$ regarding data type $m$ for a client $i$ is expressed as:
\begin{equation}\label{beta_repu}
    r_{i,m} = \mathbb{E}[Beta(a_{i,m}, b_{i,m})] = \frac{a_{i,m}+1}{a_{i,m}+b_{i,m}+2}.
\end{equation}
The update policy for the reputation model is based on accuracy improvement after the local model update from $i$, $w_{i,k,m}(t)$, is aggregated into the global model for job $k$ in the previous round, $w_{k,m}(t-1)$. If the accuracy increases, $a_{i,m}^{new} = a_{i,m} +1$; otherwise, $b_{i,m}^{new} = b_{i,m} +1$. 

The data fairness of client $i$ to job $k$ at $t$ is defined as:
\begin{equation}\label{data_fairness}
    \mathcal{F}_{i,k,m}(t) = s_{i,k,m}(t) - \frac{1}{|N_m|} \sum_{i \in N_m}s_{i,k,m}(t)
\end{equation}
where $s_{i,k,m}(t)$ is the frequency of selection for client $i$ with data type $m$ to join job $k$. $N_m$ is the set of all clients with data type $m$. If $\mathcal{F}_{i,k}(t) < 0$, it means the number of times client $i$ was selected for its dataset of type $m$ is below average. Thus, jobs shall give preference to clients with low $\mathcal{F}_{i,k,m}(t)$ values during client selection to maintain data fairness.

The reason for considering data fairness in client selection is that, if a job always only selects clients with a high reputation, this might cause oversampling of clients from specific groups. As a result, the global FL model can become biased towards the data owned by high-reputation clients. This can reduce the generalizability of FL models~\cite{Cho2020ClientSI}.

\textbf{Payment Model}: 
We adopt the Derivative Follower Pricing Strategy (DF) \cite{dasgupta2000dynamic}. DF optimizes payment efficiency. It ensures that the majority of payments are allocated towards client acquisition, while a smaller portion remains un-utilized, contributing to the earnings of FL system. It is expressed as:
\begin{equation}\label{df_pricing}
    p_k(t+1) = p_k(t) + \delta sign_1 sign_2
\end{equation}
where $sign_1 = sign(\pi_k(t)-\pi_k(t+1))$ and $sign_2 = sign(p_k(t)-p_k(t+1))$. $\pi_k(t)$ is the utility improvement in $t$. $\delta > 0$ is step size for payment change. $sign_1 sign_2 > 0$ means that the utility improvement is positively correlated to the payment value. Hence, the task publisher increases $p_k$ by one step. Otherwise, $p_k$ is decreased by one step. 

\vspace*{-1mm}
\section{The \methodname{} Approach}

\begin{algorithm}[!b]
\caption{\methodname{}\label{alg:FairFedJS_algo}}
\begin{algorithmic}[1]
\REQUIRE{Total training rounds $T$, job ordering $\mathcal{V}^t$}, FL model $w_{k,m}(t)$, selected client set $\mathcal{A}_{k,m}(t)$, data fairness $\mathcal{F}_{i,k,m}(t)$ of client $i$ for job $k$, $i$'s reputation $r_{i,m}(t)$.
\FOR{$t \leq T$}
    \STATE $\{p_1(t), \dots, p_K(t) \}$ $\leftarrow$ generate payments for each job $k$ following Eq. \eqref{df_pricing};
    \STATE Compute all $\Psi_k(t)$ following Eq. \eqref{eq:JSI};
    \STATE $\mathcal{V}^t$ $\leftarrow$ order jobs in ascending order on $\Psi_k(t)$;
    \FOR{task $k \in \mathcal{V}^t$}
        \FOR{each active client $i$}
        \STATE $\gamma_{i,k,m} = r_{i,m}(t) - \beta \mathcal{F}_{i,k,m}(t)$;
        \ENDFOR
        \STATE $\mathcal{A}_{k,m}(t)$ $\leftarrow$ select clients based on $\gamma_{i,k,m}$;
        \STATE $a_{k,m}(t) = |\mathcal{A}_{k,m}(t)|$;
    \ENDFOR
    \STATE FL training of Job $k$ with  $\mathcal{A}_{k,m}(t)$
    \STATE Update $r_{i,m}(t)$ \& $\mathcal{F}_{i,k,m}(t)$ using Eq. \eqref{beta_repu} \& Eq. \eqref{data_fairness};
    \STATE Update $Q_m(t+1)$ following Eq. \eqref{queuing_dynamics};
\ENDFOR
\end{algorithmic}
\end{algorithm}
The proposed \methodname{} approach aims to promote fair treatment of FL jobs and optimize the revenue of the FL system under the multi-job FL setting. To achieve this design goal, we leverage Lyapunov optimization to convert the dual objectives into one single objective function~\cite{meyn2012markov}.

To this end, we first introduce a virtual queue that reflects the availability of clients with dataset type $m$ at time $t$. The queuing dynamic can be expressed as:
\begin{equation}
    Q_m(t+1) = \max[0, Q_m(t) + \mu_m(t) - a_m(t)]. \label{queuing_dynamics}
\end{equation}
It reflects the net demand for clients with dataset type $m$ (i.e., the longer the queue length, the higher the demand). 
$\mu_m(t) = {\sum_{k=1}^K}{n_{k,m}}$ is the demand for clients with dataset type $m$ for this round. 
$a_m(t) = {\sum_{k=1}^K}{a_{k,m}(t)}$ is the supply of clients with dataset type $m$, where $a_{m,k}(t)$ is the number of clients with $m$ that are available to be mobilized for job $k$ in round $t$. Based on the definition of $a_m(t)$ and $\mu_m(t)$, $Q_m(t+1)$ can be converted into $Q_m(t+1) = {\sum_{k=1}^K}Q_{k,m}(t+1)$.

We leverage Lyapunov optimization to bound every increment of $Q_m(t)$. We define a quadratic Lyapunov function 
$L(\Theta(t)) = \frac{1}{2} {\sum_{m=1}^M} Q_m^2(t)$.
It is designed to measure the unfairness when allocating clients with different data types to FL jobs. A smaller value of $L(\Theta)$ is desirable as it indicates that the level of unfairness in job scheduling is low.

Then, we formulate the \textit{Lyapunov drift}, $\Delta(\Theta(t))$, to measure the expected increase of $L(\Theta(t)$ in one time step, $\Delta(\Theta(t)) = \mathbb{E}[L(\Theta(t+1))-L(\Theta(t))|\Theta(t)]$. 
By minimizing $\Delta(\Theta(t))$, we aim to limit the growth of all $Q_m(t)$ by dynamically assigning clients with multiple datasets to different types of jobs based on the dataset demand. Let $\theta = \frac{1}{2}(\mu_m^{\max})^2 + \frac{1}{2}(a_m^{\max})^2 \geq 0$, $\Delta(\Theta(t))$ is expressed as:
\begin{equation}
    \Delta(\Theta(t))\leqslant {\sum_{m=1}^M} \left(Q_m(t)[\mu_m(t) -a_m(t)] + {\theta} \right).\label{drift}
\end{equation}

Another objective is to optimize the revenue of the FL system. It is desirable for the FL system to encourage the FL jobs to increase payments for recruiting clients in order to enhance their scheduling priority.
To this end, we define the expected utility for the FL system at time $t$, $\delta(t)$, as:
\begin{equation}
    \delta(t) = \sum_{k=1}^K \frac{a_{k}(t)}{n_{k}} p_k(t) - c_k(t) \label{utility}.
\end{equation}
It reflects the income for the FL system after distributing the remuneration to the selected FL clients.

To jointly optimize the dual objectives of scheduling fairness and revenue maximization, we design a (\textit{drift} - \textit{utility}) objective function as $\Delta(\Theta(t)) - \sigma \delta(t) \label{dual_objective}$. $\sigma>0$ is a control parameter that enables system administrators to indicate their preference between these two objectives. By substituting Eq.~\eqref{drift} and Eq.~\eqref{utility} into this objective function, we have:
\begin{equation}
    \begin{aligned}
    & \frac{1}{T}{\sum_{t=0}^{T-1}}\left(\Delta (\Theta(t)) -\sigma \mathbb{E}[U(t)|\Theta(t)] \right) \geqslant \\
    & {\theta} + \frac{1}{T} {\sum_{t=0}^{T-1}}[{\sum_{m=1}^M} (Q_m(t)\mu_m(t)-Q_m(t)a_m(t)) \\
    & - \sigma \sum_{k=1}^K (\frac{a_{k}(t)}{n_{k}} p_k(t) - c_k(t))]\\
    & = {\theta} + \frac{1}{T} {\sum_{t=0}^{T-1}} \{[{\sum_{m=1}^M}{\sum_{k=1}^K} Q_{k,m}(t){n}_{k,m}(t)-Q_{k,m}(t){a_{k,m}(t)}]\\
    &- [\sigma {\sum_{k=1}^K} (\frac{\sum_{m=1}^M a_{k,m}(t)}{n_{k}} p_k(t) - \sum_{m=1}^M \frac{\hat{c}_m(t)}{\hat{r}_m(t)} a_{k,m}(t))]\}.
    \end{aligned} 
\end{equation}

As \methodname{} only controls the scheduling choices $\boldsymbol{A_t}= \{a_{1,m}(t), a_{2,m}(t), \dots, a_{K,m}(t)\}$ for each training round, we focus on the terms containing $a_{k,m}(t)$. Thus, we have
$ {\sum_{k=1}^K} [\sum_{m=1}^M -Q_{k,m}(t)a_{k,m}(t)-\sigma\frac{p_k(t)}{n_{k}} \sum_{m=1}^M a_{k,m}(t)+\sigma\sum_{m=1}^M \frac{\hat{c}_m(t)}{\hat{r}_m(t)} a_{k,m}(t)] $. 
Since we assume that each horizontal FL job requires only one type of dataset (i.e., $m=1$ for each job $k$), the objective function can be simplified as:
\begin{equation}\label{ob_fn}
    \begin{aligned}
     \min_{\boldsymbol{A_t}} \quad {\sum_{k=1}^K} a_{k}(t)\left[-Q_{k}(t) - \sigma\frac{p_k(t)}{n_{k}} + \sigma\frac{\hat{c}_m(t)}{\hat{r}_m(t)}\right].
    \end{aligned}
\end{equation}

\begin{table*}[ht]
\centering
\caption{Performance comparison results under IID and Non-IID scenarios. The best performance is indicated with \textbf{bold} text. The second best performance is indicated with \underline{underline}.}
\label{table:result_table}
\resizebox{0.68\linewidth}{!}{
\begin{tabular}{|c|c|ccc|ccc|ll}
\cline{1-8}
\multirow{2}{*}{\textbf{Method}}    & \multirow{2}{*}{\textbf{Dataset}} & \multicolumn{3}{c|}{\textbf{IID setting}}                                                                                                                            & \multicolumn{3}{c|}{\textbf{Non-IID setting}}                                                                                                                                                 & \multicolumn{1}{c}{\textbf{}} & \multicolumn{1}{c}{\textbf{}} \\ \cline{3-8}
                                    &                                   & \multicolumn{1}{c|}{\textbf{Acc(\%)}} & \multicolumn{1}{c|}{\textbf{SF}}                    & \textbf{\begin{tabular}[c]{@{}c@{}}Convergence\\ (rounds)\end{tabular}} & \multicolumn{1}{c|}{\textbf{Acc(\%)}} & \multicolumn{1}{c|}{\textbf{SF}}                    & \multicolumn{1}{c|}{\textbf{\begin{tabular}[c]{@{}c@{}}Convergence\\ (rounds)\end{tabular}}} &                               &                               \\ \cline{1-8}
\multirow{2}{*}{Random}             & Cifar-10                            & \multicolumn{1}{c|}{{\underline{57.80}}}     & \multicolumn{1}{c|}{\multirow{2}{*}{17.34}}         & \multirow{2}{*}{95.5}                                                   & \multicolumn{1}{c|}{\textbf{52.19}}  & \multicolumn{1}{c|}{\multirow{2}{*}{21.52}}         & \multirow{2}{*}{125}                                                    &                               &                               \\ \cline{2-3} \cline{6-6}
                                    & FMNIST                          & \multicolumn{1}{c|}{\textbf{86.88}}  & \multicolumn{1}{c|}{}                               &                                                                         & \multicolumn{1}{c|}{83.68}           & \multicolumn{1}{c|}{}                               &                                                                         & \multicolumn{1}{c}{}          & \multicolumn{1}{c}{}          \\ \cline{1-8}

\multirow{2}{*}{ALT}                & Cifar-10                            & \multicolumn{1}{c|}{\textbf{57.82}}  & \multicolumn{1}{c|}{\multirow{2}{*}{15.06}}         & \multirow{2}{*}{101.0}                                                  & \multicolumn{1}{c|}{{\underline{51.93}}}     & \multicolumn{1}{c|}{\multirow{2}{*}{12.23}}         & \multirow{2}{*}{98}                                                                          &                               &                               \\ \cline{2-3} \cline{6-6}
                                    & FMNIST                          & \multicolumn{1}{c|}{86.30}           & \multicolumn{1}{c|}{}                               &                                                                         & \multicolumn{1}{c|}{80.68}           & \multicolumn{1}{c|}{}                               &                                                                                              & \multicolumn{1}{c}{}          & \multicolumn{1}{c}{}          \\ \cline{1-8}
\multirow{2}{*}{UB}                 & Cifar-10                            & \multicolumn{1}{c|}{57.64}           & \multicolumn{1}{c|}{\multirow{2}{*}{15.83}}         & \multirow{2}{*}{89.2}                                                   & \multicolumn{1}{c|}{50.51}           & \multicolumn{1}{c|}{\multirow{2}{*}{14.23}}         & \multirow{2}{*}{126}                                                                         &                               &                               \\ \cline{2-3} \cline{6-6}
                                    & FMNIST                          & \multicolumn{1}{c|}{86.49}           & \multicolumn{1}{c|}{}                               &                                                                         & \multicolumn{1}{c|}{\textbf{84.00}}  & \multicolumn{1}{c|}{}                               &                                                                                              & \multicolumn{1}{c}{}          & \multicolumn{1}{c}{}          \\ \cline{1-8}
\multirow{2}{*}{MJ-FL}              & Cifar-10                            & \multicolumn{1}{c|}{57.69}           & \multicolumn{1}{c|}{\multirow{2}{*}{{\underline{5.14}}}}    & \multirow{2}{*}{{\underline{84.0}}}                                             & \multicolumn{1}{c|}{51.47}           & \multicolumn{1}{c|}{\multirow{2}{*}{{\underline{7.96}}}}    & \multirow{2}{*}{{\underline{96.7}}}                                                                  &                               &                               \\ \cline{2-3} \cline{6-6}
                                    & FMNIST                          & \multicolumn{1}{c|}{86.47}           & \multicolumn{1}{c|}{}                               &                                                                         & \multicolumn{1}{c|}{81.23}           & \multicolumn{1}{c|}{}                               &                                                                                              & \multicolumn{1}{c}{}          & \multicolumn{1}{c}{}          \\ \cline{1-8}
\multirow{2}{*}{\methodname{}} & Cifar-10                            & \multicolumn{1}{c|}{57.68}           & \multicolumn{1}{c|}{\multirow{2}{*}{\textbf{3.89}}} & \multirow{2}{*}{\textbf{83.1}}                                          & \multicolumn{1}{c|}{51.84}           & \multicolumn{1}{c|}{\multirow{2}{*}{\textbf{4.82}}} & \multirow{2}{*}{\textbf{95.8}}                                                               &                               &                               \\ \cline{2-3} \cline{6-6}
                                    & FMNIST                          & \multicolumn{1}{c|}{{\underline{86.60}}}     & \multicolumn{1}{c|}{}                               &                                                                         & \multicolumn{1}{c|}{{\underline{83.74}}}     & \multicolumn{1}{c|}{}                               &                                                                                              &                               &                               \\ \cline{1-8}
\end{tabular}
}
\end{table*}

To solve Eq.~\eqref{ob_fn}, we first propose the \textit{Job Scheduling Index} (JSI), $\Psi_k(t)$, for every individual job as:
\begin{equation}
    \Psi_k(t)= -Q_{k}(t) - \sigma\frac{p_k(t)}{n_{k}} + \sigma\frac{\hat{c}_m(t)}{\hat{r}_m(t)}.\label{eq:JSI}
\end{equation}
\methodname{} arranges FL jobs in ascending order of on their JSI values. This ranking serves the purpose of streamlining job scheduling. Higher-ranked jobs (with lower JSI values) are given priority in the client selection process. A job selects clients with consideration of both their reputation values and their data fairness values. 

 Algorithm \ref{alg:FairFedJS_algo} illustrates how \methodname{} performs job scheduling.
One advantage of \methodname{} is its ability to enable jobs to adjust their payments in order to modify their scheduling priority. When a job increases its payment at round $t$, its JSI decreases correspondingly. Consequently, this adjustment enhances its priority in the schedule.

\vspace*{-2mm}
\section{Experimental Evaluation}
We study \methodname{} through comparative experiments.

\textbf{Settings}: We conduct experiments with two public datasets: 1) Fashion-MNIST and 2) CIFAR-10. We employ diverse models to simulate various job types, setting up six concurrent jobs. Three of these jobs (i.e., MLP, CNN, ResNet) utilize the Fashion-MNIST dataset, while the remaining three (i.e., MLP, CNN, ResNet) utilize the CIFAR-10 dataset. Each job involves 10 clients.
We have a total of 50 FL clients: 20 with Fashion-MNIST, 20 with CIFAR-10, and the remaining 10 with both datasets. Images are randomly sampled from training sets and distributed to clients, each receiving 1,400 samples. 
To simulate different data quality levels, 
we investigate two experimental scenarios with IID data and Non-IID data, following the data settings in~\cite{shi2023fairness}.
In addition, we assign random costs, denoted as $c_{i,m} \in [1,3]$, for each dataset $m$ belonging to client $i$. For job payments, we initialize them randomly from $[10, 12, \dots, 28, 30]$, updating them in each round according to the proposed payment update policy in Eq. \eqref{df_pricing}, with a step size of $\delta=2$.

\textbf{Comparison Baselines}: We compare \methodname{} with 4 existing approaches: 
1) \textbf{Random (Rand)}: In each round, the job selection order is scheduled randomly. 
2) \textbf{Alternating (ALT)}: The job selection order in the current round is the reverse of that in the previous round.
3) \textbf{Utility-Based (UB)}: The job selection order in the current round is based on their utility in the previous round. Jobs with lower utility in the previous round are more eager to boost their selection order in the current round.
4) \textbf{MJ-FL}~\cite{zhou2022efficient}: In the original context, MJ-FL (BODS-based) optimizes the scheduling policy based on its cost function. In our context, we adapt it to use client reputation value instead of time cost.

\textbf{Evaluation Metrics}: The performance of the comparison approaches is assessed based on convergence time (in terms of round number) and test accuracy. Moreover, we evaluate job scheduling fairness (SF) by examining the variance of queue lengths as 
$SF= \sqrt{\frac{{\sum_{t=1}^T}{\sum_{m=1}^M} (Q_m(t)- \overline{Q}(t))^2}{T} }$.
$\overline{Q}(t)$ is the average queue length at round $T$. $SF$ measures the long-term variations in all queue lengths. A higher $SF$ indicates that the demands for certain dataset types are not promptly fulfilled, leading to extended waiting time for jobs requiring such data. 
We closely monitor SF when the demand for clients is higher than the supply.
A high $SF$ value implies low fairness.
\textbf{Results Discussion}: Table~\ref{table:result_table} shows the average performance of comparison approaches. 
The results show that \methodname{} significantly outperforms all the baselines in terms of scheduling fairness ($SF$). It achieves 31.9\% lower $SF$ on average than the best-performing baseline MJ-FL. 
This demonstrates that our key design of incorporating the availability queue length of client datasets into job scheduling criteria effectively enhances scheduling fairness. 
\methodname{} achieves comparable test accuracy with the baselines, and beats the best baseline MJ-FL by 1.0\% in terms of convergence time.


\vspace*{-2mm}
\section{Conclusions and Future Work}
In this paper, we propose a fairness-aware job scheduling approach \methodname{} for multi-job FL. By leveraging Lyapunov optimization, it ensures fair allocation of high-demand FL client datasets to FL jobs by considering both current demand and job payment bids, preventing prolonged waiting. Extensive experiments highlight that \methodname{} outperforms the best baselines with an average improvement of 31.9\% in scheduling fairness and 1.0\% in convergence time while achieving comparable test accuracy. It makes the most advantageous trade-offs among test accuracy, convergence rate and fairness compared to the state-of-the-art approaches. To the best of our knowledge, it is the first fairness-aware job scheduling approach designed for multi-job FL.

In subsequent research, we plan to adapt \methodname{} to vertical FL \cite{Li-et-al:2023INFOCOM} and federated transfer learning \cite{Feng-et-al:2022} scenarios where each job may require multiple types of data.

\bibliographystyle{IEEEbib}
\bibliography{strings}

\end{document}